\documentclass[letterpaper, 10 pt, conference]{ieeeconf}  % Comment this line out if you need a4paper

\IEEEoverridecommandlockouts                              % This command is only needed if 
\usepackage{fix-cm} % Allow the 7.5 pt table to use matching math font sizes.
\usepackage{amsmath}
\usepackage{amssymb}
\usepackage{booktabs}
\usepackage{graphicx}
\usepackage{float}
\usepackage[table]{xcolor}
\definecolor{citeblue}{RGB}{150, 84, 226}

\makeatletter
\let\NAT@parse\undefined
\makeatother

\usepackage[
    breaklinks=true,
    colorlinks=true,
    linkcolor=citeblue,
    urlcolor=citeblue,
    citecolor=citeblue,
    bookmarks=false
]{hyperref}

\usepackage[all]{hypcap}
\definecolor{bestyellow}{HTML}{FFE9AD}
\definecolor{mopapurple}{HTML}{9654e2}
\definecolor{mopaorange}{HTML}{f4952f}

\newcommand{\method}{MoPA}
\newcommand{\titlemethod}{\textcolor{mopapurple}{Mo}\textcolor{mopaorange}{PA}}
\newcommand{\mstd}[2]{\ensuremath{#1\,{\color{gray}\scriptstyle\pm #2}}}
\DeclareRobustCommand{\bestmark}{%
  {\setlength{\fboxsep}{1pt}\colorbox{bestyellow}{\textbf{Best}}}%
}

\title{\LARGE \bf
\titlemethod{}: Coordinated \textcolor{mopapurple}{Mo}bile Manipulation via\\
Subsystem-Specific \textcolor{mopaorange}{P}erception \textcolor{mopaorange}{A}lignment
}

\author{Guangyu Chen$^{1,2,*}$, Qiwei Liang$^{2,3,*}$, Shaolong Zhu$^{1,2,*}$,
Tianxing Chen$^{2,4,*}$, Zikuan Xiao$^{2}$,\\
Yifan Xie$^{1}$, Lingfeng Zhang$^{1}$, Ping Luo$^{4}$,
Renjing Xu$^{3,\dagger}$, and Wenbo Ding$^{1,2,\dagger}$%
\thanks{$^{*}$Equal contribution and shared first authorship.}%
\thanks{$^{\dagger}$Corresponding author: Wenbo Ding.}%
\thanks{$^{1}$THU; $^{2}$Xspark AI; $^{3}$HKUST (GZ); $^{4}$HKU.}
\\
\normalsize{\url{https://MoPA-policy.github.io}}
}

\begin{document}

\maketitle
\thispagestyle{empty}
\pagestyle{empty}

%%%%%%%%%%%%%%%%%%%%%%%%%%%%%%%%%%%%%%%%%%%%%%%%%%%%%%%%%%%%%%%%%%%%%%%%%%%%%%%%
\begin{abstract}
Mobile manipulation requires perceptual evidence at different spatial scales
for base motion and arm control, while the two action modalities remain
kinematically coupled. Existing policies often employ specialized action
generation for different subsystems but condition heterogeneous action
branches on a shared perceptual representation, leaving subsystem-specific
perception--action correspondence implicit. We present \method{}, a framework
that aligns perceptual conditioning with mobility and manipulation while
preserving coordination at the action level. Dual Perceptual Streams employ
two mutually masked query banks to extract separate perceptual representations
from a shared vision--language context. Perception2Action Adaptation jointly
updates each query bank and its corresponding action stream at every layer of
a structured Mixture-of-Transformers decoder, while enabling information
exchange between the two action streams. Coupled conditional flow matching
learns a joint vector field for coordinated generation of both action chunks.
On the ManiSkill-HAB benchmark, \method{} achieves state-of-the-art
performance across all three task suites. Across four real-world tasks,
\method{} achieves a mean full-task success rate of 76.3\%, outperforming the
best baseline by 12.5 percentage points. Ablation studies and further analyses
validate the effectiveness of the proposed design.
\end{abstract}

%%%%%%%%%%%%%%%%%%%%%%%%%%%%%%%%%%%%%%%%%%%%%%%%%%%%%%%%%%%%%%%%%%%%%%%%%%%%%%%%
\section{INTRODUCTION}

Robots deployed in unstructured household environments must move between
workspaces and interact with objects as their egocentric views continuously
change. Unlike fixed-base tabletop manipulation, mobile manipulation requires
joint control of a mobile base and a manipulator. The base relies on
scene-level geometry for repositioning, whereas the manipulator requires local
geometry and interaction state for precise end-effector control. Their action
spaces differ, yet their motions are coupled through a single kinematic chain:
base-pose errors propagate to the end effector, while arm reachability
constrains feasible base motion. A policy must therefore organize perceptual
evidence by subsystem and generate coordinated whole-body actions
~\cite{mobilealoha,harmonicmm,spin}.

Recent work improves perception under changing viewpoints by combining
multiview observations with geometric, temporal, and semantic context
~\cite{dspv2,acdit,incom,sbp,geohat,chen2025g3flow,chen2026rmbench}. Complementary work structures action
generation for heterogeneous control through hierarchical or parallel
decoders, token routing, and specialized Transformer streams
~\cite{mobileumi,mobilewam,abotm05}. Together, these directions expand the
evidence available to a policy and tailor action computation to heterogeneous
control, yet the action branches generally remain conditioned on a shared or
globally fused representation.

\begin{figure}[t]
    \centering
    \includegraphics[width=\columnwidth]{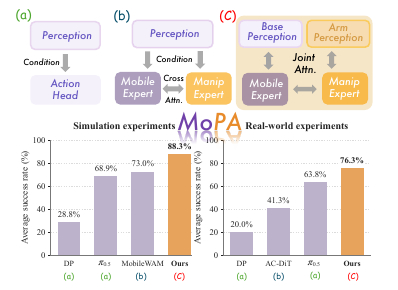}
    \caption{Mobile-manipulation paradigms and performance.
    (a) Shared perception conditions a unified action head.
    (b) Separate base and manipulation experts share perceptual conditioning.
    (c) \method{} combines subsystem-specific perception with action-level
    coordination. Bottom: average
    success rates in SetTable simulation and real-world experiments.}
    \label{fig:teaser}
    \vspace{-1em}
\end{figure}

Despite these advances, subsystem-level perception--action alignment remains
unresolved. Mobile manipulation requires selecting perceptual evidence
according to each subsystem's control objective and generating actions through
subsystem-specific pathways while preserving coordination~\cite{liang2026whole}. Existing methods
typically address these requirements separately, linking perception and action
through a global representation. During simultaneous base and arm motion, that
representation must encode both scene layout and local interaction geometry,
despite the different spatial scales emphasized by the two subsystems. Even
with separate action heads, losses from both branches can update shared
projection or fusion layers, potentially inducing gradient conflicts in
multi-task optimization~\cite{recon}. Separating output parameters therefore
does not guarantee that the action experts receive subsystem-specific
perceptual conditioning.

Existing perception--action interfaces also struggle to support this
separation. Directly supplying fused features to multiple action branches gives
all experts the same, undifferentiated conditioning. Allowing action tokens to
retrieve information independently from the full perceptual sequence requires
repeated access to a large token set and provides no persistent intermediate
representation aligned with a specific action space. We therefore argue that
mobile manipulation requires a mechanism that bridges shared perception and
heterogeneous action experts. Such a mechanism should derive subsystem-specific
representations from a common vision--language context, update each
representation as its corresponding actions are generated, and retain
coordination between the base and manipulator at the action level.

As summarized in \hyperref[fig:teaser]{Fig.~\ref*{fig:teaser}}, we introduce \method{}, a
subsystem-aligned mobile-manipulation framework with two complementary modules.
\emph{Dual Perceptual Streams} connect a shared Qwen3-VL
context~\cite{qwen3vl} to heterogeneous action experts. Given the same head-
and wrist-camera RGB images and language instruction, isolated \emph{Manip
Query} and \emph{Mobile Query} banks aggregate subsystem-relevant evidence.
A structured mask prevents direct interaction between the banks, yielding
distinct perceptual representations before action decoding.

\emph{Perception2Action Adaptation} adapts these perceptual representations to
heterogeneous action generation while coordinating whole-body control. A
structured Mixture-of-Transformers decoder assigns separate action tokens and
parameter sets to manipulation and mobility. Each query bank interacts
bidirectionally only with its matched action stream, while the two action
streams exchange their evolving control states. Layerwise query--action updates
transform perceptual summaries into action-conditioned states. We use coupled
flow matching to train the model and jointly sample the two action chunks.
Dual Perceptual Streams thus provide separate perceptual conditioning for the
two subsystems, while Perception2Action Adaptation adapts and coordinates their
action generation.

On ManiSkill-HAB~\cite{mshab}, \method{} achieves the highest macro-averaged
skill success rates among the compared methods on SetTable
(\textbf{88.3\%}), TidyHouse (\textbf{72.2\%}), and PrepareGroceries
(\textbf{67.8\%}). These results exceed the strongest baseline on each suite
by 4.5, 17.0, and 7.1 percentage points, respectively.

Our contributions are:
\begin{itemize}
    \item We characterize the lack of explicit subsystem-level
    perception--action correspondence and explain why separating action heads
    alone is insufficient.
    \item We propose \emph{Dual Perceptual Streams}, which use isolated
    \emph{Manip Query} and \emph{Mobile Query} banks to derive subsystem-specific
    perceptual representations from a shared vision--language context.
    \item We propose \emph{Perception2Action Adaptation}, which jointly updates
    these representations with their matched action tokens and coordinates
    separate action streams.
    \item We achieve state-of-the-art performance on all three
    ManiSkill-HAB task suites and validate our design through extensive
    ablation studies and exploratory experiments.
\end{itemize}

\section{Related Work}

\textbf{Visuomotor Policies.}
Mobile manipulation policies map onboard observations and robot state
to whole-body commands. Mobile ALOHA and HarmonicMM jointly model
mobile manipulation, while SPIN adds active camera
control~\cite{mobilealoha,harmonicmm,spin}.
CausalMoMa introduces causal credit assignment, while VBC and UMI on
Legs decompose control into body/base targets and task-frame
end-effector trajectories~\cite{causalmoma,vbc,umionlegs}.
DSPv2, AC-DiT, and InCoM refine multiview semantic--geometric
features; SBP maintains a persistent 3D map, and GeoHAT routes
reliable visual tokens to base and arm
actions~\cite{dspv2,acdit,incom,sbp,geohat}.
Mobile UMI factorizes manipulation and base
trajectories~\cite{mobileumi}.
These methods improve grounding and execution, but typically select
subsystem-specific evidence within the action decoder rather than
constructing persistent perceptual conditions before decoding.

\textbf{Vision--Language--Action Policies.}
VLA policies use pretrained vision--language priors for
instruction-conditioned control.
$\pi_{0.5}$ leverages heterogeneous training for long-horizon tasks.
MoManipVLA transfers fixed-base policies through base-waypoint and
end-effector trajectory optimization, while Mobi-$\pi$ searches at
deployment for an in-distribution base pose for an existing
policy~\cite{pi05,momanipvla,mobipi}.
G0 and G0.5 integrate planning, reasoning, and action
tokenization~\cite{g0,g05}.
SG-VLA, EchoVLA, SERF, and PanoVLA enrich context through multiview
observations, memory, maps, or panoramic views; AnchorVLA and
MolmoB0T address efficient closed-loop generation and sim-to-real
transfer, respectively~\cite{sgvla,echovla,serf,panovla,anchorvla,molmobot}.
Nevertheless, whole-body policies generally share an action
interface, leaving perception--action correspondence for mobility
and manipulation implicit.

\textbf{World--Action Models.}
WAMs augment observations with predicted visual or latent dynamics.
DreamZero jointly predicts video and actions, MotionWAM uses
video-model denoising features for control, and $\omega$-0 conditions
diffusion actions on future-observation
embeddings~\cite{dreamzero,motionwam,omega0}.
For mobile manipulation, MobileWAM routes computation among shared,
mobility, and manipulation experts; ABot-M0.5 combines latent
actions with a Mixture-of-Transformers; and DECOWAM separates base
and arm latents~\cite{mobilewam,abotm05,decowam}.
World--Ego Modeling factorizes world/ego prediction, while
DreamTrajectory ranks candidate whole-body
trajectories~\cite{wem,dreamtrajectory}.
These methods specialize mainly through future latents or action
experts; \method{} instead establishes an explicit, persistent
interface between shared perceptual context and coordinated
action streams.

\section{Method}
\label{sec:method}

\begin{figure*}[!t]
    \centering
    \includegraphics[width=\textwidth]{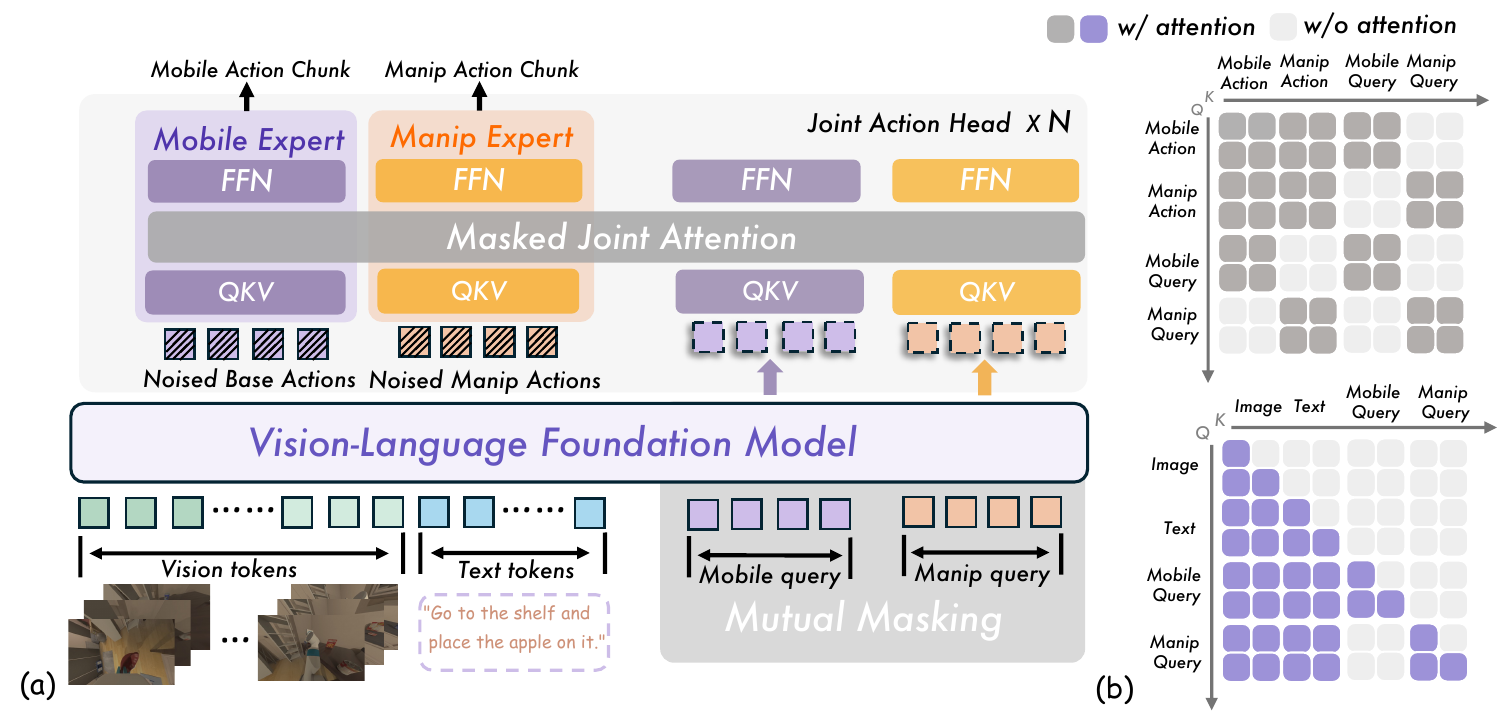}
    \caption{Overview of \method{}. (a) A shared
    vision--language foundation model produces separate \emph{Mobile Query}
    and \emph{Manip Query} streams that condition dedicated Base
    and Manipulation experts in a joint action head. (b) Structured visibility
    patterns in the VLM and action head: direct attention between the query
    streams is masked, each query stream connects to its matched action stream,
    and the two action streams remain coupled for whole-body coordination.}
    \label{fig:main}
\end{figure*}

This section presents \method{}, a subsystem-aligned perception--action
framework for mobile manipulation. We first define the observation and action
spaces (Sec.~\ref{sec:problem_formulation}) and outline the architecture
(Sec.~\ref{sec:method_overview}). We then introduce Dual Perceptual Streams,
which derive branch-specific perceptual representations from shared
vision--language tokens (Sec.~\ref{sec:dual_query}), and Perception2Action
Adaptation, which models and
coordinates the two action streams (Sec.~\ref{sec:structured_dmot}). Finally,
we describe the coupled conditional flow-matching objective and synchronous
inference procedure (Sec.~\ref{sec:flow_matching}).

\subsection{Problem Formulation}
\label{sec:problem_formulation}

At time $t$, the policy receives head- and wrist-camera RGB images
$I_t^{\mathrm{head}}$ and $I_t^{\mathrm{wrist}}$, a proprioceptive state
$\mathbf{s}_t\in\mathbb{R}^{d_s}$, and a language instruction $\ell$.
Following the input/output convention of AC-DiT~\cite{acdit}, it predicts a
whole-body action chunk of horizon $H$,
$\mathbf{A}_t=(\mathbf{A}_t^{\mathrm{man}},\mathbf{A}_t^{\mathrm{mob}})$.
For branch $k\in\{\mathrm{man},\mathrm{mob}\}$,
$\mathbf{A}_t^k=(\mathbf{a}_t^k,\ldots,\mathbf{a}_{t+H-1}^k)
\in\mathbb{R}^{H\times d_k}$. The policy models the conditional distribution
\begin{equation}
    p_{\Theta}\!\left(
    \mathbf{A}_t^{\mathrm{man}},
    \mathbf{A}_t^{\mathrm{mob}}
    \mid I_t^{\mathrm{head}},I_t^{\mathrm{wrist}},
    \mathbf{s}_t,\ell\right).
    \label{eq:joint_policy}
\end{equation}
The manipulation and mobility branches control the arm/gripper and torso/base,
respectively. Joint modeling allows the policy to learn their coordination
directly from whole-body demonstrations.

\subsection{Method Overview}
\label{sec:method_overview}

\method{} comprises three stages. First, Qwen3-VL encodes head- and
wrist-view images and the language instruction into shared
vision--language tokens, from which two isolated query banks,
\emph{Manip Query} and \emph{Mobile Query}, extract branch-specific
perceptual streams. Proprioception bypasses the VLM and is encoded
separately in each action branch. Next, Perception2Action Adaptation
uses a structured Mixture-of-Transformers (MoT) decoder to pair each
perceptual stream with its corresponding action stream while allowing
the two action streams to exchange information. Finally, coupled
conditional flow matching learns a joint vector field to generate
both action chunks synchronously. Dual Perceptual Streams determine
each subsystem's perceptual conditioning, while Perception2Action
Adaptation models their actions and coordination.
\hyperref[fig:main]{Fig.~\ref*{fig:main}} shows the architecture and
information flow.

\subsection{Dual Perceptual Streams}
\label{sec:dual_query}

The shared vision--language context contains scene-level cues for mobility and
local interaction cues for manipulation. When both action experts receive the
same representation, each must extract the information relevant to its control
objective from an undifferentiated conditioning signal. Dual Perceptual Streams
construct a dedicated perceptual representation for each subsystem before action
decoding.

Let $\mathbf{P}_t$ denote the shared vision--language token sequence, and let
$\mathbf{Q}^{\mathrm{Manip}}$ and $\mathbf{Q}^{\mathrm{Mob}}$ be the learned
Manip Query and Mobile Query banks. We append both banks to $\mathbf{P}_t$:
\begin{equation}
    \mathbf{Z}_t =
    \left[\mathbf{P}_t,
    \mathbf{Q}^{\mathrm{Manip}},
    \mathbf{Q}^{\mathrm{Mob}}\right].
    \label{eq:query_sequence}
\end{equation}
Let $\mathbf{H}_t^{\mathrm{Manip}}$ and $\mathbf{H}_t^{\mathrm{Mob}}$ denote
the final VLM hidden states at the positions of the two query banks. As shown
in the lower matrix of \hyperref[fig:main]{Fig.~\ref*{fig:main}(b)}, both banks attend to
$\mathbf{P}_t$, while direct attention between them is masked. They therefore
aggregate the same context through separate paths. Distinct learned embeddings,
separate aggregation paths, and supervision from the corresponding action
branches encourage each bank to retain information relevant to its control
objective. The resulting states are routed only to their matched action
streams, providing distinct perceptual representations before action prediction.
Perception2Action Adaptation then handles cross-subsystem coordination.

\subsection{Perception2Action Adaptation}
\label{sec:structured_dmot}

Subsystem-specific perceptual representations alone do not ensure coordinated
whole-body control: independently decoded manipulation and mobility commands
can still conflict. Perception2Action Adaptation addresses this issue with a
structured MoT decoder that uses separate parameters for the two action
branches while coupling the evolving states of both action streams.

Separate projectors first map the perceptual states into the feature space of
the action model:
\begin{equation}
    \overline{\mathbf{Q}}_{t,0}^q =
    \Phi^q\!\left(\mathbf{H}_t^q\right),
    \qquad q\in\{\mathrm{Manip},\mathrm{Mob}\}.
    \label{eq:query_projection}
\end{equation}
For each branch $k\in\{\mathrm{man},\mathrm{mob}\}$, we concatenate the
proprioceptive and action tokens at flow time $\tau$ to form the initial
branch sequence:
\begin{equation}
    \mathbf{X}_{\tau,0}^k =
    \left[
        E_s^k(\mathbf{s}_t)+\mathbf{e}_s^k,\,
        E_a^k(\mathbf{A}_{\tau}^k,\tau)
        +\mathbf{E}_{\mathrm{pos}}^k
    \right].
    \label{eq:action_stream}
\end{equation}
The projectors $\Phi^q$ and encoders $E_s^k,E_a^k$ are branch-specific;
$\mathbf{e}_s^k$ and $\mathbf{E}_{\mathrm{pos}}^k$ are the state-token type
embedding and action-token positional embeddings, respectively.
Here, $q=\mathrm{Manip}$ and $q=\mathrm{Mob}$ index the query streams, while
$k=\mathrm{man}$ and $k=\mathrm{mob}$ index their matched action branches.
The continuous flow time is represented by a timestep embedding in the action
encoder and in each decoder block.

The joint action head processes the two query streams and two action streams
using the structured connectivity in the upper matrix of \hyperref[fig:main]{Fig.~\ref*{fig:main}(b)}.
Each query bank interacts with its corresponding action stream, while the two
action streams exchange information to coordinate whole-body control.

Let $L$ denote the number of decoder blocks. For $l=0,\ldots,L-1$, the $l$-th
block, represented by $\mathcal{B}_l$, performs the joint update
\begin{align}
    &\left(
    \overline{\mathbf{Q}}_{t,l+1}^{\mathrm{Manip}},
    \overline{\mathbf{Q}}_{t,l+1}^{\mathrm{Mob}},
    \mathbf{X}_{\tau,l+1}^{\mathrm{man}},
    \mathbf{X}_{\tau,l+1}^{\mathrm{mob}}
    \right) \nonumber\\
    &\quad =
    \mathcal{B}_l\!\left(
    \overline{\mathbf{Q}}_{t,l}^{\mathrm{Manip}},
    \overline{\mathbf{Q}}_{t,l}^{\mathrm{Mob}},
    \mathbf{X}_{\tau,l}^{\mathrm{man}},
    \mathbf{X}_{\tau,l}^{\mathrm{mob}};
    \tau
    \right).
    \label{eq:layer_update}
\end{align}
Each stream has its own projection, normalization, and feed-forward parameters.
At each velocity-field evaluation, the query states are initialized using
Eq.~\eqref{eq:query_projection} and updated jointly with the action tokens at
every layer. They thus evolve from perceptual summaries into action-conditioned
control states within that evaluation. The final action tokens are decoded
into flow velocities for manipulation and mobility.

\subsection{Coupled Conditional Flow Matching}
\label{sec:flow_matching}

We train the coupled decoder to model a conditional vector field over the
joint whole-body action space. For each branch $k$, let
$\mathbf{A}_1^k\equiv\mathbf{A}_t^k$ denote a clean demonstration chunk and
sample an independent noise chunk $\mathbf{A}_0^k\in\mathbb{R}^{H\times d_k}$
with i.i.d. standard Gaussian entries. The clean chunks of both branches come
from the same demonstration interval, preserving their temporal
correspondence across the horizon. At a flow time
$\tau\sim\mathcal{U}(0,1)$ shared by both branches, the linear interpolation
and its constant target velocity are
\begin{equation}
    \mathbf{A}_{\tau}^k =
    (1-\tau)\mathbf{A}_0^k+\tau\mathbf{A}_1^k,
    \qquad
    \mathbf{U}^k=\mathbf{A}_1^k-\mathbf{A}_0^k.
    \label{eq:flow_path}
\end{equation}
The flow time $\tau$ parameterizes the transformation from noise to actions,
whereas $t$ indexes interaction with the environment. The target velocity
$\mathbf{U}^k$ is therefore defined in action space over the entire chunk.

Conditioned on the projected query states and encoded branch sequences, the
joint action head $G_{\psi}$ predicts both branch velocities:
\begin{align}
    &\left(
    \widehat{\mathbf{U}}^{\mathrm{man}},
    \widehat{\mathbf{U}}^{\mathrm{mob}}\right) \nonumber\\
    &\quad =
    G_{\psi}\!\left(
    \overline{\mathbf{Q}}_{t,0}^{\mathrm{Manip}},
    \overline{\mathbf{Q}}_{t,0}^{\mathrm{Mob}},
    \mathbf{X}_{\tau,0}^{\mathrm{man}},
    \mathbf{X}_{\tau,0}^{\mathrm{mob}};
    \tau
    \right).
    \label{eq:velocity_prediction}
\end{align}
Here, $\psi$ denotes the action-head parameters, and
$\widehat{\mathbf{U}}^k\in\mathbb{R}^{H\times d_k}$.
Because the action streams exchange information in every decoder block, each
prediction can depend on the current state of the other branch. Although the
target in Eq.~\eqref{eq:flow_path} is constant for a sampled
noise--demonstration pair, the predicted field depends on the interpolated
actions, perceptual conditioning, and flow time. We optimize the joint vector
field with a branch-normalized flow-matching objective:
\begin{align}
    \mathcal{L}_{\mathrm{FM}}
    = \mathbb{E}_{\mathbf{A}_1,\mathbf{A}_0,\tau}\!\left[
    \sum_{k\in\{\mathrm{man},\mathrm{mob}\}}
    \frac{\lambda_k}{H d_k}
    \left\|
    \widehat{\mathbf{U}}^k-\mathbf{U}^k
    \right\|_F^2
    \right],
    \label{eq:flow_loss}
\end{align}
where $\lambda_k$ balances the branch contributions, and the squared
Frobenius norm sums the squared errors over the $H\times d_k$ entries.
Normalization by $H d_k$ prevents the higher-dimensional branch from
dominating the objective solely because it has more action channels.
The vision--language backbone, Dual Perceptual Streams, and
Perception2Action Adaptation are optimized end to end with
Eq.~\eqref{eq:flow_loss}.

% Results are discussed in Sec.~\ref{sec:sim_benchmark}.
\begin{table*}[!t]
    \caption{SetTable benchmark results.
    Entries report mean success rates (\%), with standard deviations in gray.
    For each metric, rankings among evaluated methods are marked as
    \bestmark{} and \textbf{Second Best}.
    $^{*}$ denotes depth input.}
    \label{tab:settable_breakdown}
    \centering
    \fontsize{7.5pt}{9pt}\selectfont
    \setlength{\tabcolsep}{3pt}
    \begin{tabular*}{\textwidth}{@{\extracolsep{\fill}}lcccccccc}
        \toprule
        Method & Pick Apple & Pick Bowl & Place Apple & Place Bowl & Open Fridge & Open Drawer & Close Drawer & Mean \\
        \midrule
        DP3$^{*}$~\cite{dp3} & \mstd{0.0}{0.0} & \mstd{20.0}{2.4} & \mstd{31.0}{0.8} & \mstd{32.0}{0.8} & \mstd{0.0}{0.0} & \mstd{0.0}{0.0} & \mstd{68.0}{0.0} & 21.6 \\
        ACT~\cite{act} & \mstd{28.0}{2.2} & \mstd{28.0}{2.4} & \mstd{8.7}{3.3} & \mstd{13.0}{0.8} & \mstd{2.0}{2.2} & \mstd{0.0}{0.0} & \mstd{85.7}{1.2} & 23.6 \\
        DP~\cite{diffusionpolicy} & \mstd{21.3}{3.3} & \mstd{20.7}{3.3} & \mstd{28.0}{8.0} & \mstd{69.3}{3.3} & \mstd{7.3}{5.8} & \mstd{0.0}{0.0} & \mstd{55.0}{5.7} & 28.8 \\
        RDT-1B~\cite{rdt1b} & \mstd{12.0}{11.3} & \mstd{10.7}{6.8} & \mstd{32.0}{5.7} & \mstd{18.7}{5.0} & \mstd{82.7}{10.5} & \mstd{44.0}{8.6} & \cellcolor{bestyellow}\mstd{\mathbf{100.0}}{0.0} & 42.9 \\
        AC-DiT$^{*}$~\cite{acdit} & \mstd{33.3}{1.9} & \mstd{36.0}{6.5} & \mstd{33.3}{9.4} & \mstd{17.3}{6.8} & \mstd{90.7}{5.0} & \mstd{81.3}{6.8} & \mstd{\mathbf{97.3}}{1.9} & 55.6 \\
        $\pi_0$~\cite{pi0} & \mstd{26.6}{3.8} & \mstd{26.6}{4.5} & \mstd{48.9}{5.2} & \mstd{56.8}{5.9} & \mstd{90.5}{1.9} & \mstd{75.5}{2.5} & \mstd{88.7}{2.0} & 59.1 \\
        AnchorVLA~\cite{anchorvla} & \mstd{22.7}{0.9} & \mstd{44.5}{0.8} & \mstd{64.3}{0.8} & \mstd{63.8}{2.0} & \mstd{88.9}{0.8} & -- & \cellcolor{bestyellow}\mstd{\mathbf{100.0}}{0.0} & 64.0 \\
        $\pi_{0.5}$~\cite{pi05} & \mstd{44.3}{2.3} & \mstd{45.3}{2.1} & \mstd{57.3}{2.9} & \mstd{65.7}{1.2} & \mstd{92.3}{0.6} & \mstd{86.7}{3.5} & \mstd{90.7}{1.5} & 68.9 \\
        MobileWAM~\cite{mobilewam} & \mstd{46.0}{0.8} & \mstd{46.0}{2.6} & \mstd{63.7}{3.2} & \mstd{64.7}{1.2} & \cellcolor{bestyellow}\mstd{\mathbf{99.3}}{0.5} & \mstd{\mathbf{91.0}}{0.8} & \cellcolor{bestyellow}\mstd{\mathbf{100.0}}{0.0} & 73.0 \\
        GeoHAT$^{*}$~\cite{geohat} & \mstd{\mathbf{82.3}}{2.5} & \mstd{69.3}{1.3} & \mstd{60.0}{1.6} & \mstd{78.0}{0.0} & \mstd{83.7}{0.5} & \mstd{86.0}{0.0} & \mstd{95.3}{0.5} & 79.3 \\
        InCoM$^{*}$~\cite{incom} & \mstd{59.4}{4.3} & \mstd{\mathbf{84.1}}{4.6} & \cellcolor{bestyellow}\mstd{\mathbf{84.1}}{5.4} & \mstd{\mathbf{82.5}}{5.6} & \mstd{87.3}{2.8} & \mstd{88.9}{2.3} & \cellcolor{bestyellow}\mstd{\mathbf{100.0}}{0.0} & \textbf{83.8} \\
        \midrule
        \textbf{\method{} (Ours)} & \cellcolor{bestyellow}\mstd{\mathbf{84.3}}{0.6} & \cellcolor{bestyellow}\mstd{\mathbf{89.0}}{2.0} & \mstd{\mathbf{72.7}}{0.6} & \cellcolor{bestyellow}\mstd{\mathbf{83.7}}{2.5} & \mstd{\mathbf{98.3}}{0.6} & \cellcolor{bestyellow}\mstd{\mathbf{94.0}}{1.0} & \mstd{96.3}{0.6} & \cellcolor{bestyellow}\textbf{88.3} \\
        \bottomrule
    \end{tabular*}
\end{table*}

At inference, both action chunks start from independent Gaussian noise and
are integrated synchronously using $N$ Euler steps of size $1/N$.
At each shared flow time $\tau_n=n/N$, for $n=0,\ldots,N-1$, the decoder
uses the current pair of chunks to predict velocities
$\widehat{\mathbf{U}}_n^k$, and the updates are
\begin{equation}
    \mathbf{A}_{\tau_{n+1}}^k =
    \mathbf{A}_{\tau_n}^k
    +\frac{1}{N}\widehat{\mathbf{U}}_n^k,
    \qquad
    k\in\{\mathrm{man},\mathrm{mob}\}.
    \label{eq:euler_sampling}
\end{equation}
Both velocities are evaluated from the same pre-update pair of action chunks
before either update is applied. The visual--language inputs and
proprioceptive state remain fixed throughout these $N$ steps. At
$\tau_N=1$, we concatenate the resulting manipulation and mobility chunks
along the action dimension to form the whole-body chunk. The robot executes
its initial actions and replans from updated observations.

\section{Experiments}
\label{sec:experiments}

We evaluate \method{} on simulation benchmarks
(Sec.~\ref{sec:sim_benchmark}), examine its core components through ablations
(Sec.~\ref{sec:ablations}), explore query capacity and representational differentiation
(Sec.~\ref{sec:analysis}), and assess real-world performance
(Sec.~\ref{sec:real_robot}).

\subsection{Simulation Experiments}
\label{sec:sim_benchmark}

\noindent\textbf{Experimental Setup.}
We evaluate \method{} on ManiSkill-HAB~\cite{mshab}, a household
mobile-manipulation benchmark built on ManiSkill3 and SAPIEN. It combines
ReplicaCAD apartment scenes with YCB object models and supports whole-body
control of a Fetch mobile manipulator. Our evaluation covers three suites:
SetTable, TidyHouse, and PrepareGroceries. For SetTable, we evaluate seven
skills: picking and placing an apple and a bowl, opening a fridge, and
opening and closing a kitchen drawer. TidyHouse involves object rearrangement
across open receptacles, whereas PrepareGroceries involves transferring
objects between a fridge and kitchen counters. Each of these two suites
contains 18 pick/place skills across nine object categories. Together, the
suites require coordinated base and arm motion in cluttered scenes for both
object rearrangement and interaction with articulated objects.

\noindent\textbf{Implementation Details.}
We train one policy per suite using 1,000 successful demonstrations per
skill, with head- and wrist-camera RGB images, proprioception, and language
instructions as inputs~\cite{acdit}. The action chunk length is $H=4$, with an 11-D
whole-body action per time step. Training uses AdamW for 100k optimization
steps with cosine learning-rate decay and a batch size of 64.
At inference, each chunk is sampled with four Euler integration steps;
the first two actions are executed at 20~Hz before replanning from
updated observations.
We evaluate each skill in three runs of 100 held-out episodes each,
with at most 200 steps per episode, and report the mean and standard
deviation of success rates across runs. Each suite score is the
macro-average of its skill success rates.

\noindent\textbf{Baselines.}
We compare representative end-to-end imitation-learning methods evaluated
on ManiSkill-HAB across three paradigms: \emph{Visuomotor Policies},
including DP3~\cite{dp3}, ACT~\cite{act}, DP~\cite{diffusionpolicy}, DSPv2~\cite{dspv2},
AC-DiT~\cite{acdit}, GeoHAT~\cite{geohat}, InCoM~\cite{incom}, and
RDT-1B~\cite{rdt1b}; \emph{Vision--Language--Action Policies}, including
$\pi_0$~\cite{pi0}, $\pi_{0.5}$~\cite{pi05}, and AnchorVLA~\cite{anchorvla};
and \emph{World--Action Models}, represented by
MobileWAM~\cite{mobilewam}. Methods that use depth information as input
are marked with $^{*}$ in \hyperref[tab:settable_breakdown]{Table~\ref*{tab:settable_breakdown}}.

% Declare Fig. 3 before Fig. 4 to preserve figure numbering.
\begin{figure*}[!t]
    \centering
    \includegraphics[width=\textwidth]{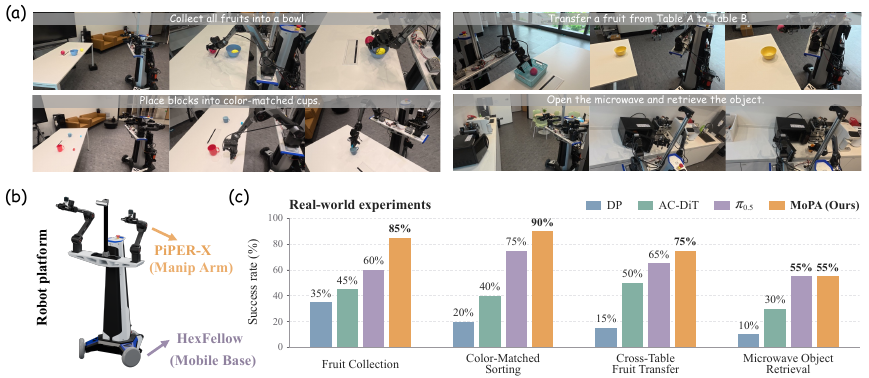}
    \caption{Real-world evaluation of \method{}. (a) Execution sequences
    for fruit collection, color-matched sorting, cross-table fruit transfer,
    and microwave object retrieval. (b) Robot platform. (c) Success rates
    compared with DP, AC-DiT, and $\pi_{0.5}$.}
    \label{fig:real_exp}
\end{figure*}

\noindent\textbf{Results and Analysis.}
\method{} achieves the highest macro-averaged skill success rate among
the compared methods across all three suites. On SetTable, it achieves
88.3\%, surpassing InCoM by 4.5 percentage points and leading on four
of seven skills
(\hyperref[tab:settable_breakdown]{Table~\ref*{tab:settable_breakdown}}).
On TidyHouse and PrepareGroceries, it achieves 72.2\% and 67.8\%,
respectively, outperforming the strongest baseline, $\pi_{0.5}$
(55.2\% and 60.7\%), by 17.0 and 7.1 percentage points
(\hyperref[tab:mshab_summary]{Table~\ref*{tab:mshab_summary}}).
Gains over $\pi_{0.5}$ cover both picking and placement on these two
suites, with a larger improvement on TidyHouse's Pick than Place
category (22.5 vs.\ 11.5 percentage points). InCoM nevertheless
retains the lead on SetTable's Place Apple and TidyHouse's Place category.

Related baselines employ stage-adaptive perception~\cite{incom},
geometry-aware visual selection~\cite{geohat}, or specialized
action experts~\cite{mobilewam}. On SetTable, MobileWAM achieves
99.3\% on Open Fridge and 100.0\% on Close Drawer, but only 46.0\%
on each picking task, compared with \method{}'s 84.3\% on Pick Apple
and 89.0\% on Pick Bowl. This gap motivates aligning perceptual
conditioning with each subsystem alongside action specialization.

\begin{figure}[h]
    \centering
    \includegraphics[width=\columnwidth]{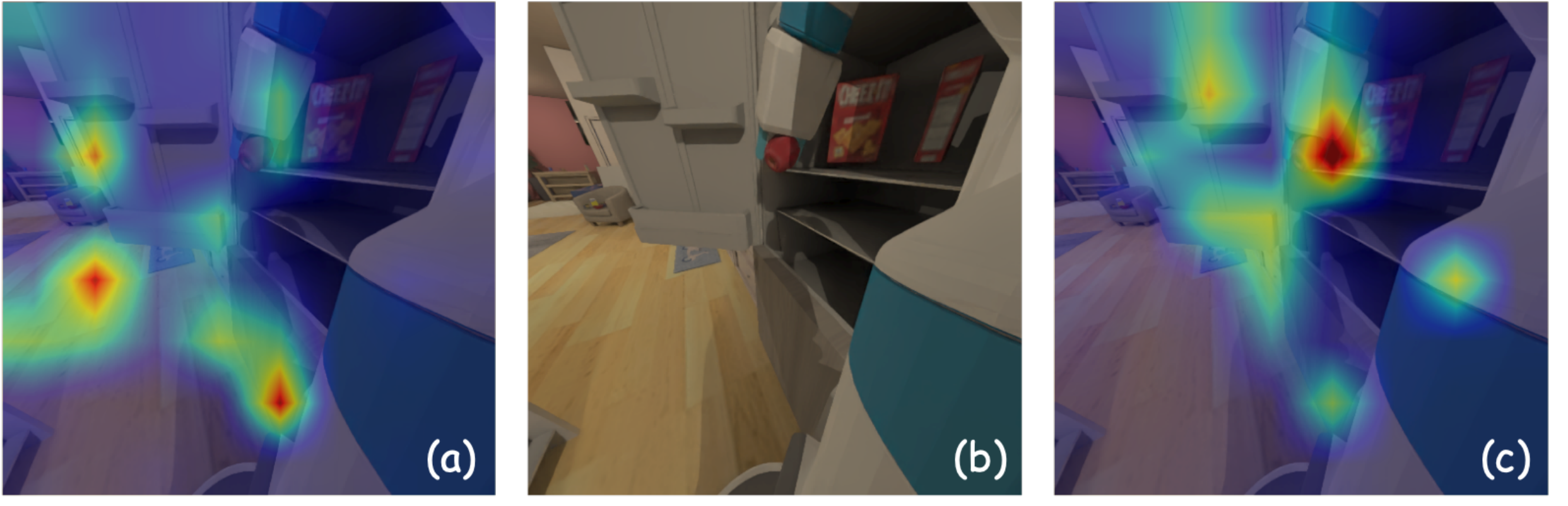}
    \caption{Query attention visualization. (a) Mobile Query attention,
    (b) the RGB observation, and (c) Manip Query attention.}
    \label{fig:query_attention}
\end{figure}

\method{} constructs separate perceptual streams and refines them
during coordinated action generation. In the example shown in
\hyperref[fig:query_attention]{Fig.~\ref*{fig:query_attention}},
Mobile Query emphasizes the floor and surrounding scene structure,
while Manip Query focuses on the shelf around the target object,
consistent with their intended roles. Performance drops under
Shared Query, w/o Joint Attention, and w/o Corresponding Query Access
(\hyperref[tab:component_ablation]{Table~\ref*{tab:component_ablation}})
support subsystem-specific conditioning and its adaptation during
action decoding.

\begin{table}[h]
    \caption{Mean success rates (\%) on TidyHouse and PrepareGroceries.
    Standard deviations are gray; \bestmark{} and \textbf{Second Best}
    are marked per column.}
    \label{tab:mshab_summary}
    \centering
    \scriptsize
    \setlength{\tabcolsep}{2.2pt}
    \begin{tabular*}{\columnwidth}{@{\extracolsep{\fill}}lcccccc}
        \toprule
        & \multicolumn{3}{c}{TidyHouse}
        & \multicolumn{3}{c}{PrepareGroceries} \\
        \cmidrule(lr){2-4}\cmidrule(lr){5-7}
        Method & Pick & Place & Mean & Pick & Place & Mean \\
        \midrule
        ACT~\cite{act}                    & \mstd{2.2}{0.8} & \mstd{31.6}{3.2} & 16.9 & \mstd{2.0}{1.1} & \mstd{27.5}{2.9} & 14.8 \\
        DP~\cite{diffusionpolicy}         & \mstd{0.0}{0.0} & \mstd{30.3}{3.0} & 15.2 & \mstd{0.4}{0.3} & \mstd{17.7}{1.9} & 9.1 \\
        DP3~\cite{dp3}                    & \mstd{0.0}{0.0} & \mstd{61.0}{0.0} & 30.5 & \mstd{0.0}{0.0} & \mstd{31.3}{1.2} & 15.7 \\
        DSPv2~\cite{dspv2}                & \mstd{1.3}{0.3} & \mstd{42.1}{3.1} & 21.7 & \mstd{0.9}{0.6} & \mstd{32.8}{2.9} & 16.9 \\
        InCoM~\cite{incom}                & \mstd{16.7}{1.9} & \cellcolor{bestyellow}\mstd{\mathbf{78.9}}{2.9} & 47.8 & \mstd{15.0}{2.4} & \mstd{\mathbf{65.9}}{3.1} & 40.5 \\
        GeoHAT~\cite{geohat}              & \mstd{30.3}{0.5} & \mstd{\mathbf{73.3}}{1.7} & 51.8 & \mstd{19.7}{4.6} & \mstd{60.7}{2.5} & 40.2 \\
        $\pi_{0.5}$~\cite{pi05}            & \mstd{\mathbf{51.4}}{1.3} & \mstd{58.9}{2.4} & \textbf{55.2} & \mstd{\mathbf{59.3}}{2.1} & \mstd{62.1}{3.5} & \textbf{60.7} \\
        \midrule
        \textbf{\method{} (Ours)}                & \cellcolor{bestyellow}\mstd{\mathbf{73.9}}{0.8} & \mstd{70.4}{2.3} & \cellcolor{bestyellow}\textbf{72.2} & \cellcolor{bestyellow}\mstd{\mathbf{66.6}}{1.2} & \cellcolor{bestyellow}\mstd{\mathbf{69.0}}{1.6} & \cellcolor{bestyellow}\textbf{67.8} \\
        \bottomrule
    \end{tabular*}
\end{table}

\subsection{Ablation Studies}
\label{sec:ablations}

We evaluate Dual Perceptual Streams and Perception2Action Adaptation on
SetTable (\hyperref[tab:component_ablation]{Table~\ref*{tab:component_ablation}}). We first examine how perceptual
representations are constructed, then how they are updated and matched to action
branches. All variants use the same backbone and training and evaluation
protocols, and retain both action experts and interaction between their streams.

\paragraph{Perceptual conditioning}
We first compare direct VLM conditioning with learned query aggregation.
Direct VLM Conditioning replaces queries with VLM hidden states while
retaining branch-specific projections and joint condition--action updates.
Shared Query supplies one learned query bank to both branches, increasing
the success rate from 58.4\% to 74.7\%. Separating the queries by subsystem
further raises the success rate to 88.3\% without changing the total learned
query capacity. These comparisons support both learnable extraction of
perceptual information and subsystem-specific conditioning; query aggregation
alone does not explain the overall improvement.

\paragraph{Joint query--action adaptation}
With subsystem-specific representations in place, we next examine whether they
benefit from action-dependent updates. The w/o Joint Attention variant keeps
query parameters trainable and preserves matched query access, but fixes query
hidden states within each action-head evaluation. The success rate drops to
70.3\%, 18.0 percentage points below that of the full model. Since both action
experts and their interaction are retained, this result supports adapting
perceptual representations to evolving actions rather than supplying static
summaries.

\paragraph{Subsystem-aligned access}
We next examine which query states each action stream should access during
joint updating. The w/o Corresponding Query Access variant retains both query
banks and joint updates but allows each action stream to read both banks.
Unlike Shared Query, it changes action-side access while preserving separate
query aggregation. The success rate drops to 66.3\%, 22.0 percentage points
below that of the full model. This result supports maintaining explicit
query--action correspondence during decoding, beyond constructing separate
query banks and updating them jointly with actions.

\begin{table}[!t]
    \caption{Component ablations and query bank size analysis on SetTable simulation experiments.}
    \label{tab:component_ablation}
    \centering
    \small
    \setlength{\tabcolsep}{4.5pt}
    \renewcommand{\arraystretch}{1.15}
    \begin{tabular*}{\columnwidth}{@{\extracolsep{\fill}}lcc}
        \toprule
        Variant & Avg. SR & $\Delta\mathrm{SR}$ \\
        \midrule
        Direct VLM Conditioning & 58.4 & $\downarrow\,29.9$ \\
        Shared Query & 74.7 & $\downarrow\,13.6$ \\
        w/o Joint Attention & 70.3 & $\downarrow\,18.0$ \\
        w/o Corresponding Query Access & 66.3 & $\downarrow\,22.0$ \\
        \midrule
        \method{} w/ Queries per bank $=4$ & 74.6 & $\downarrow\,13.7$ \\
        \method{} w/ Queries per bank $=16$ & 82.5 & $\downarrow\,5.8$ \\
        \midrule
        \method{} w/ Queries per bank $=8$ & \textbf{88.3} & -- \\
        \bottomrule
    \end{tabular*}
\end{table}

\subsection{Exploratory Experiments}
\label{sec:analysis}

We investigate two research questions (RQs):
\begin{itemize}
    \item \textbf{RQ1 (Query Bank Size):} Does using eight queries per action
    branch yield the highest success rate among the tested configurations?
    \item \textbf{RQ2 (Query Differentiation):} Do the Mobile Query and
    Manip Query banks exhibit distinct representations during inference?
\end{itemize}

\paragraph{RQ1: Effect of Query Bank Size}
We vary the number of queries per bank, $q\in\{4,8,16\}$, corresponding to
8, 16, and 32 total queries, while keeping the remaining architecture and
training settings fixed. The final three rows of
\hyperref[tab:component_ablation]{Table~\ref*{tab:component_ablation}} report average success rates on SetTable.

Average success increases from 74.6\% at $q=4$ to 88.3\% at $q=8$, then
decreases to 82.5\% at $q=16$. We therefore adopt the eight-query configuration,
which achieves the highest success rate among the tested settings. This sweep
assesses sensitivity to query capacity, whereas Shared Query tests the value
of subsystem differentiation with a fixed learned query budget.

\begin{figure}[H]
    \centering
    \includegraphics[width=\columnwidth]{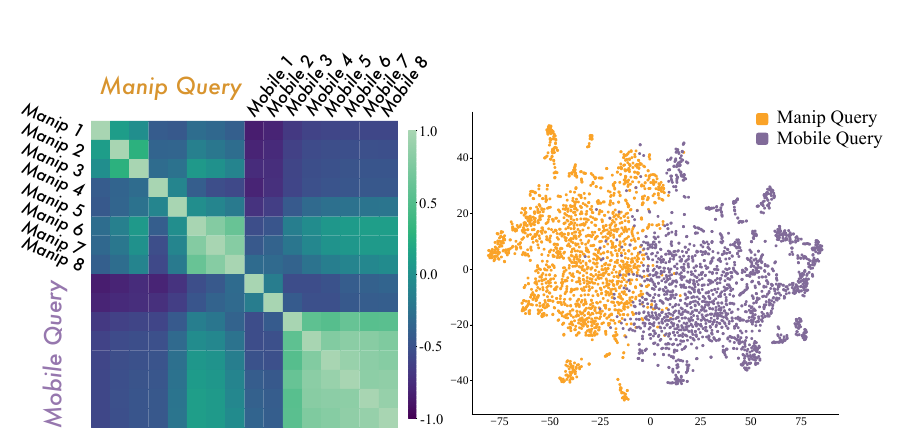}
    \caption{Query representations during inference. Left: cosine-similarity
    heatmap. Right: t-SNE projection, with orange and purple denoting
    Manip Query and Mobile Query states, respectively.}
    \label{fig:query_tsne}
\end{figure}

\paragraph{RQ2: Query Representation Differentiation}
To investigate the representational differences between Manip Queries and
Mobile Queries, we collect query states during inference on a 20\% subset
of the dataset and visualize them using a cosine-similarity heatmap and
t-SNE (\hyperref[fig:query_tsne]{Fig.~\ref*{fig:query_tsne}}). The heatmap shows low or negative cosine
similarity for many cross-type query pairs, indicating differences in their
feature directions. Consistently, the t-SNE projection shows that Manip Query
and Mobile Query states predominantly occupy distinct regions, with partial
overlap. Together, these observations provide qualitative evidence of
differentiated representations between the two query types during inference,
consistent with their intended roles in manipulation and mobility.

\vspace{-5em}

\subsection{Real-World Experiments}
\label{sec:real_robot}

\noindent\textbf{Experimental Setup.}
Our platform (\hyperref[fig:real_exp]{Fig.~\ref*{fig:real_exp}(b)})
comprises a HexFellow Trigger-A3 omnidirectional mobile base, an IOTA
vertical lift, two AgileX PiPER-X six-DoF arms, and three RGB-D cameras
(one head-mounted and one on each wrist). Policy evaluation and deployment
use XPolicyLab~\cite{community2026xpolicylab}.

We evaluate four tasks, illustrated in
\hyperref[fig:real_exp]{Fig.~\ref*{fig:real_exp}(a)}:
\begin{itemize}
    \item \emph{Fruit Collection:} Approach a table and place all fruit
    into a bowl.
    \item \emph{Color-Matched Sorting:} Approach a table and place blocks
    into color-matched cups.
    \item \emph{Cross-Table Fruit Transfer:} Pick up a fruit at Table A,
    navigate to Table B, and place it there.
    \item \emph{Microwave Object Retrieval:} Approach a microwave,
    open its door, and retrieve an object.
\end{itemize}
These tasks involve repeated pick-and-place, color-based
object--receptacle matching, and multi-stage coordination across
workspaces or during articulated-object interaction.
We collect 200 demonstrations per task.

\noindent\textbf{Implementation Details.}
We retain the simulation action chunk length $H=4$. Each time step uses
a 17-D whole-body action, including base velocities
$[v_x,v_y,\omega_z]$ for planar translation and yaw rotation.
We compare \method{} with Diffusion Policy (DP)~\cite{diffusionpolicy},
AC-DiT~\cite{acdit}, and $\pi_{0.5}$~\cite{pi05} on all four tasks.
\method{}, AC-DiT, and $\pi_{0.5}$ each use a single policy jointly
trained on all tasks for 50k optimization steps with a batch size of 64.
DP uses a separate policy per task, trained for 100 epochs with a
batch size of 128. We evaluate each method over 20 trials per task
and report full-task success rates.

\noindent\textbf{Results and Analysis.}
\hyperref[fig:real_exp]{Figure~\ref*{fig:real_exp}(c)} reports task success rates. \method{} achieves
success rates of 85.0\%, 90.0\%, 75.0\%, and 55.0\% on Fruit Collection,
Color-Matched Sorting, Cross-Table Fruit Transfer, and Microwave Object
Retrieval, respectively. Its average success rate of 76.3\% exceeds that of
the strongest baseline, $\pi_{0.5}$, by 12.5 percentage points. \method{}
achieves the highest success rate on the first three tasks and ties with
$\pi_{0.5}$ on microwave retrieval.

Fruit Collection and Color-Matched Sorting require repeated object selection
and precise placement, whereas Cross-Table Fruit Transfer combines grasping
and placement with movement between workspaces. Dual Perceptual Streams
provides separate perceptual conditions for mobility and manipulation, and
Perception2Action Adaptation refines these conditions alongside action
generation while coordinating the two action streams. The high success rates
on these tasks demonstrate that \method{} can combine fine-grained interaction
and cross-workspace execution within a single multi-task policy.

\section{Conclusion}
\label{sec:conclusion}

We presented \method{}, a mobile-manipulation framework that connects shared
vision--language representations to coordinated action generation through
subsystem-aligned perceptual streams. Dual Perceptual Streams construct
distinct perceptual representations for mobility and manipulation, while
Perception2Action Adaptation refines them jointly with their corresponding action
streams and preserves communication between the action streams. Experiments
on three ManiSkill-HAB suites and four real-world tasks show higher average
success rates than the evaluated baselines. Ablation studies and analyses
support the complementary roles of separate perceptual conditioning,
action-dependent query updates, and matched query access. These findings
support explicit perception--action correspondence as a useful design principle
for coordinated mobile manipulation. Performance remains task-dependent, as
illustrated by the tie with the strongest evaluated baseline on microwave
retrieval, motivating further evaluation across more diverse environments and
interaction demands.

\bibliographystyle{IEEEtran}
\bibliography{references}

\end{document}